\documentclass{article}
\usepackage{spconf,amsmath,graphicx}
\usepackage{comment}
\usepackage{multirow}
\usepackage[noadjust]{cite}

\title{Fast and accurate factorized neural transducer for text adaption of end-to-end speech recognition models}
\name{Rui Zhao, Jian Xue, Partha Parthasarathy,Veljko Miljanic, Jinyu Li}
\address{Microsoft Speech Group, Redmond, WA, USA}
\begin{document}
\ninept
\maketitle
\begin{abstract}
Neural transducer is now the most popular end-to-end model for speech recognition, due to its naturally streaming ability. However, it is challenging to adapt it with text-only data. Factorized neural transducer (FNT) model was proposed to mitigate this problem. The improved adaptation ability of FNT on text-only adaptation data came at the cost of lowered accuracy compared to the standard neural transducer model. We propose several methods to improve the performance of the FNT model. They are: adding CTC criterion during training, adding KL divergence loss during adaptation, using a pre-trained language model to seed the vocabulary predictor, and an efficient adaptation approach by interpolating the vocabulary predictor with the n-gram language model. A combination of these approaches results in a relative word-error-rate reduction of 9.48\% from the standard FNT model. Furthermore, n-gram interpolation with  the vocabulary predictor improves the adaptation speed hugely with satisfactory adaptation performance. 
\end{abstract}
\begin{keywords}
neural transducer model, factorized transducer model, KL divergence, n-gram 
\end{keywords}
\section{Introduction}
\label{sec:intro}

Recently, neural transducer based end-to-end (E2E) models \cite{Graves-RNNSeqTransduction, he2019streaming, attentionisallyouneed, yeh2019transformer, TT, Li2019RNNT, battenberg2017exploring,chiu2018state, Li2020comparison,  xiechen, E2EOverview}, such as recurrent neural network transducer (RNN-T) \cite{Graves-RNNSeqTransduction} , transformer-transducer (T-T) \cite{yeh2019transformer, TT} and conformer-transducer (C-T) \cite{gulati2020conformer}, have become the dominant model for automatic speech recognition (ASR) in industry  due to its natural streaming property, as well as competitive accuracy with traditional hybrid speech recognition systems \cite{watanabe2017hybrid, sainath2020streaming, Li2020Developing}. 

However, one of the main challenges for neural transducer models is adaptation using only text data. This is because in neural transducer models, there are no separated acoustic or language model like in traditional hybrid models. Although the prediction network could be considered as an internal language model (LM) because the input to it is the previously predicted token, it is not a real LM since the prediction output needs to be combined with the acoustic encoder in a non-linear way to generate  posteriors over the vocabulary augmented with a \textit{blank} token. Adapting the prediction network using text-only data is not as straightforward or effective as adapting the LM in hybrid systems. Paired audio and text data is needed to adapt a neural transducer model, however,  collecting labeled audio data is both time and money costly. 

There are several types of methods proposed to address this issue. One is to generate artificial audio for adaptation text instead of collecting real audio. Audio generation method could be based on multi-speaker neural text to speech (TTS) model \cite{Li2020Developing, sim2019personalization, deng2020ttsrnnt, zheng2021ttsasr, ttsjasha} or spliced data method \cite{spliced}. The neural transducer model could then be fine-tuned with artificial paired audio and text data. A major drawback of these kinds of methods is the high computational cost. It takes much longer for the TTS-based methods to generate audio  even with GPU machines, while the spliced-data method  has very small cost for generating audio. However, the adaptation step for both methods involves updating part of the encoder, full prediction and the joint network with the RNN-T loss. The result is high computational cost for training, need for GPUs, and too much delay for scenarios in which rapid adaptation is necessary. 

Another class of text-only adaptation methods is LM fusion \cite{kannan2018shallowfusion, 2020fusion, 2021fusion, amazonilm, triebiasing}, such as shallow fusion \cite{kannan2018shallowfusion} where an external LM trained on target-domain text is incorporated during the neural transducer model decoding. However, there is already an internal LM in the neural transducer model. Directly adding an external LM is not mathematically grounded. To solve such an issue,  density ratio \cite{mcdermott2019densityratio}, hybrid autoregressive transducer model \cite{variani2020hybrid}, and internal LM estimation \cite{meng2021ilme,ibmilm} were proposed to remove the influence of the internal LM contained in the neural transducer model. However, the performance is often sensitive to the interpolation weight of external LM for different tasks, and it needs to be well tuned based on development data to get optimal results \cite{meng202ilmt}. 

Different from aforementioned methods, factorized neural transducer model (FNT) \cite{fnt} modifies the neural transducer model architecture by factorizing the blank and vocabulary prediction so that a standalone LM can be used for the vocabulary prediction. In this way, various language model adaptation [32, 33, 34] techniques could be applied to FNT. But based on results in \cite{fnt}, FNT degrades the accuracy on general testing sets compared with the standard neural transducer model, although it significantly improves the accuracy in the new domain after adaptation. Besides, it still needs significant GPU time to finetune FNT with text only data for the adaptation, which may not meet the fast adaptation requirement in some real applications. 

In this paper, we proposed several methods to advance FNT for effective and efficient adaptation with text only data. These methods include: 1) Adding Connectionist Temporal Classification (CTC) \cite{Graves-CTCFirst} criterion for the encoder network during training to make it work more like an acoustic model.  Then, the combination of encoder output and vocabulary predictor output is similar to the combination of acoustic and language model in hybrid models.  2) Adding Kullback-Leibler (KL) divergence between the outputs of adapted model and baseline model to avoid over fitting to the adaptation data. 3) Initializing the vocabulary predictor with a neural LM trained with more text data. 4) Replacing the network fine-tuning with more efficient adaptation method using n-gram interpolation. Experimental results showed that on general testing sets, these methods help the modified FNT to get even a little better accuracy than the baseline neural transducer model. On adaptation sets, the word error rate (WER) after the adaptation of modified FNT is reduced by 9.48\% relatively from the standard FNT model, and reduced by 29.21\% relatively from the baseline C-T model. Besides, n-gram interpolation results in much faster adaptation speed. 

The rest of this paper is organized as follows: Section \ref{sec:rnnt} introduces the neural transducer model and FNT model. Section \ref{sec:refinefnt} presents the proposed methods for modified FNT in detail. Section \ref{sec:exp} shows the experimental results and analysis. Section \ref{sec:conclusion} gives the conclusions.


\section{standard neural transducer and FNT model}
\label{sec:rnnt}
\subsection{Standard neural transducer}
\label{ssec:strnnt}

A neural transducer model \cite{Graves-RNNSeqTransduction} consists of encoder, prediction, and joint networks. The encoder network is analogous to the acoustic model in hybrid models, which converts the acoustic feature $x_t$ into a high-level representation $f_t$, where $t$ is the time index. The prediction network works like a neural LM, which produces a high-level representation $g_u$ by conditioning on the previous non-blank target $y_{u-1}$ predicted by the RNN-T model, where $u$ is output label index.  The joint network combines the encoder network output $f_t$ and the prediction network output $g_u$ to compute the output probability with  
\begin{eqnarray}
z_{t, u} = W *\text{relu}(f_t+g_u)+b \nonumber \\
 P(\hat{y}_{t+1}|x_1^t, y_1^{u}) = \text{softmax}(z_{t, u}) 
\end{eqnarray}

To address the length differences between the acoustic feature $\textbf{x}_1^T$ and label sequences $\textbf{y}_1^U$,
a special blank symbol, $\phi$, is added to the output vocabulary. Therefore the output set is $\{{\phi} \cup \mathcal{V}\}$, where $\mathcal{V}$ is the vocabulary set. 

\subsection{Factorized neural transducer}
\label{ssec:fnt}

\begin{figure}[htb]
\centering
\includegraphics[width=5.5cm]{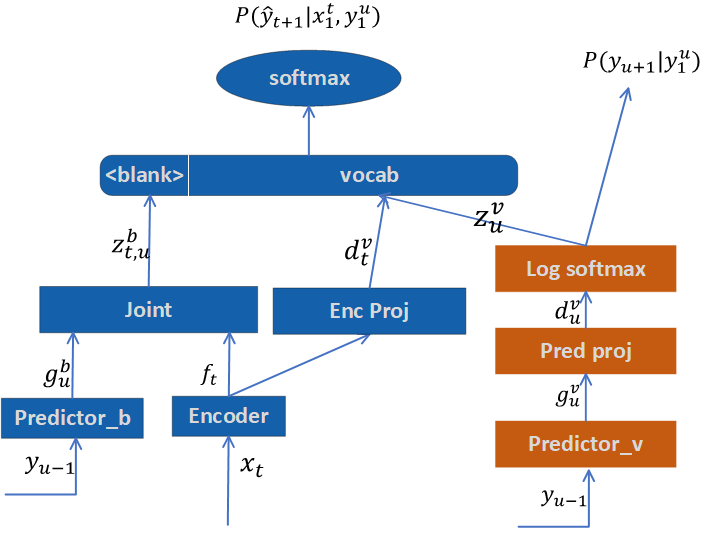}
\caption{Flowchart of factorized neural transducer}
\label{fig:fnt}
 \vspace{-0.2cm}
\end{figure}

Two prediction networks are used in FNT \cite{fnt}, as shown in figure \ref{fig:fnt}. One ($Predictor\_b$) is for the prediction of the blank label $\phi$,
and the other ($Predictor\_v$) is vocabulary prediction (rightmost orange part in figure \ref{fig:fnt}). The vocabulary predictor could be considered as a standard LM. 
The combination methods with encoder output $f_t$ for these two prediction outputs are different. For the blank prediction, it is the same as in standard neural transducer models. 
\begin{align}
z^{b}_{t, u} = W^b *\text{relu}(f_t+g_u^b)+b^b
\end{align}
For the vocabulary prediction, it is firstly projected to the vocabulary size and converted to the log probability domain by the operation of log softmax. After this, it is added with the encoder output. 
\begin{eqnarray}
    d_t^v &=& W_{enc}^v * \text{relu}(f_t)+b_{enc}^v \nonumber \\
    d_u^v &=& W_{pred}^v* \text{relu}(g_u^v)+b_{pred}^v \nonumber \\
    z_u^v &=& \text{log\_softmax}(d_u^v) \nonumber \\
    z^{v}_{t, u} &=& d_t^v + z_u^v 
\label{eqn:fnt}
\end{eqnarray}
Two combination outputs are concatenated and softmax is applied to get the final label probability
\begin{align}
P(\hat{y}_{t+1}|x_1^t, y_1^{u}) = \text{softmax}([z^b_{t, u};  z^{v}_{t,u}] )
\end{align}
The loss function of FNT is
\begin{equation}
\mathcal{J}_f = \mathcal{J}_t - \lambda \log P(\textbf{y}_1^U)
\label{eqn:4}
\end{equation}
where the first term is the standard neural transducer loss and the second term is the LM loss with cross entropy (CE). 
$\lambda$ is a hyper-parameter to tune the effect of LM loss. 


\section{Improvement of FNT}
\label{sec:refinefnt}

In this section, we will  propose several methods to improve the accuracy and efficiency of FNT. 

\subsection{Adding CTC criterion in training}
\label{ssec:ctc}

As showed in section \ref{ssec:fnt}, the encoder output and the vocabulary predictor output are combined by sum operation. The predictor output is log probability, but the encoder output is not. According to Bayes' theory, the acoustic and language model scores should be combined by  weighted sum in log probability domain. Therefore we refine FNT by converting the encoder output to the log probability by adding log softmax 

\begin{figure}[htb]
\centering
\includegraphics[width=8cm]{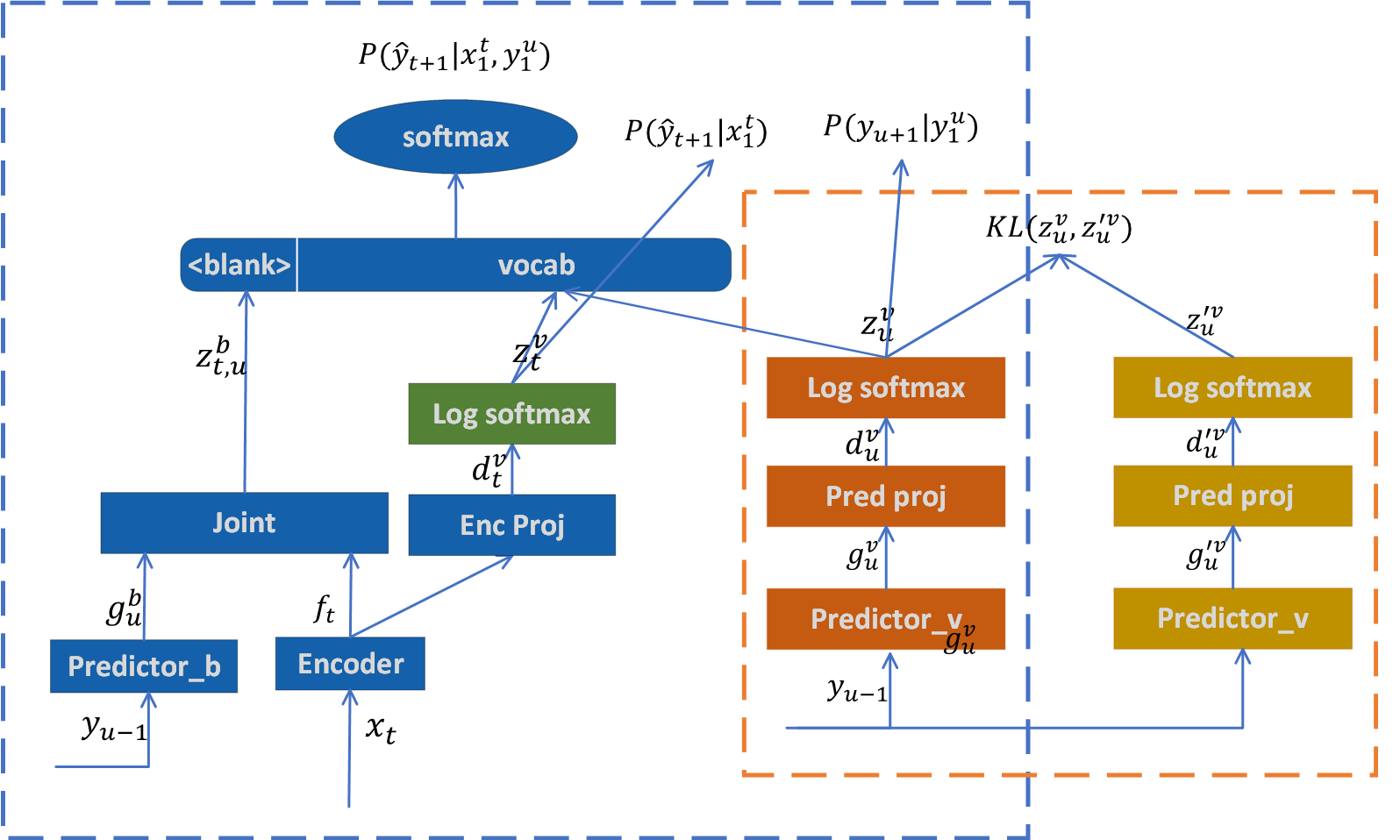}
\caption{BLUE FRAME: adding CTC criterion for FNT training. ORANGE FRAME: adding KL divergence loss for FNT adaptation }
\label{fig:fnt_refine}
 \vspace{-0.2cm}
\end{figure}

Furthermore, to force the encoder part act more like the acoustic model, CTC criterion is added for the encoder output as shown in the blue frame part in figure \ref{fig:fnt_refine}. The reason we choose CTC instead CE is that it's not easy to get the sentence piece level alignment for training data, while sentence piece unit is commonly used as the output unit for neural transducer E2E model. 

With such changes, the combination of encoder output and vocabulary predictor output is shown in below equations. 

\begin{eqnarray}
    z_t^v &=& \text{log\_softmax}(d_t^v) \nonumber \\
    z^{v}_{t, u} &=& z_t^v[:-1] + \gamma * z_u^v  \label{eq:merge}
\label{eqn:ctc}
\end{eqnarray}
where $\gamma$ is a trainable parameter, which could be taken as LM weight. One thing needs to be mentioned is after adding CTC, the dimension of $d_t^v$ and $z_t^v$ become vocabulary\_size+1 because CTC needs one extra output ``blank''. Here we put ``blank'' as the last dimension. and it's excluded when $z_t^v$ is added with $z_u^v$.  

The final loss function can be written as 
\begin{equation}
\mathcal{J}_f = \mathcal{J}_t - \lambda \log P(\textbf{y}_1^U)+\beta \mathcal{J}_{ctc}
\label{eqn:loss}
\end{equation}
where $\mathcal{J}_{ctc}$ is CTC loss and $\beta$ is a hyper-parameter to be tuned in the experiments.

\subsection{Adding KL divergence in adaptation}
\label{ssec:kl}
To adapt FNT model with text data, the most straightforward way is to finetune the vocabulary predictor with the adaptation text based on cross-entropy loss. But this may degrade model performance on general domain. To avoid this, KL divergence between the vocabulary predictor outputs of adapted model and baseline model is added during the adaptation as shown in the orange frame part in figure \ref{fig:fnt_refine}. 

The adaptation loss with KL divergence is
\begin{equation}
\mathcal{J}_{adapt} = \text{CE}(Z_u^v,Y_{adapt}) + \alpha \text{KL}(Z_u^v,Z_u^{'v})
\label{eqn:kl}
\end{equation}
where $Y_{adapt}$ is the adaptation text data, $Z_u^{v}$ is the log softmax of adaptation text from the adapted model and $Z_u^{'v}$ is the log softmax of adaptation text from the baseline model. $\alpha$ is the KL divergence weight to be tuned in the experiments.

\subsection{External language model}
\label{ssec:lm}

Since the vocabulary predictor in FNT is designed to be an LM, we explore the possibility of training it independently on a much larger text corpus than the transcriptions in the FNT training data. The parameters
of this pre-trained external LM could be further updated to potentially improve accuracy. In principle, we could choose a variety of architectures for the external LM. In this paper, we limit ourselves to an architecture that is very close to that of the standard prediction network for a fair comparison with the baseline system. The vocabulary of the external LM is the same as that of the FNT system, and is trained using the conventional cross-entropy loss. 
Experimental results in Section \ref{sec:exp} show that external LM trained with more data improves model accuracy, and updating the external LM parameters during FNT model training further improves the results.

\subsection{N-gram interpolation}
\label{ssec:ngram}

As noted before, fine-tuning the vocabulary predictor is one of the straightforward adaptation method for FNT. Although updating vocabulary predictor is much faster than updating the whole FNT network, it could not meet the immediate adaptation requirement for some applications. In this paper, we propose to use n-gram integration for fast adaptation of FNT.

In this method, a n-gram based language model is firstly trained with the adaptation text data. Then it is interpolated with the probability output from the vocabulary predictor during the decoding. The vocabulary log probability after interpolation with n-gram probability $P(y_u|y_1^{u-1})_{ngram}$ is calculated as 
\begin{align}
z_u^v = \text{log}((1-w)*P(y_u|y_1^{u-1})_{pred}+w*P(y_u|y_1^{u-1})_{ngram})
\label{eqn:ngram}
\end{align}
where $P(y_u|y_1^{u-1})_{pred} = \text{softmax}(d_u^v)$ is the label probability output from vocabulary predictor. Then $z_u^v$ is plugged into Equation \eqref{eq:merge} to calculate $z^{v}_{t, u}$ which is used to generate the final output of FNT.

In this method, no neural network training is involved, and n-gram LM model training is super fast with the adaptation text data. Experimental results in section \ref{ssec:result_ngram} show it has much lower computational cost compared to the fine-tuning based method.


\section{experiments}
\label{sec:exp}

In this section, the effectiveness of the proposed methods are evaluated based on conformer-transducer (C-T) model \cite{gulati2020conformer} for several adaptation tasks with different amount of adaptation text data. 

In the baseline C-T model, the encoder network contains 18 conformer layers. The prediction network contains 2 LSTM \cite{hochreiter1997long} layers, and 1024 nodes per layer. The output label size is 4000 sentence pieces. We use the low-latency streaming implementation in \cite{xiechen} with 160 milliseconds (ms)  encoder lookahead. 
The standard FNT model has the same encoder structure and output label inventory as the baseline C-T model. The blank and vocabulary predictor each consists of 2 LSTM layers, also 1024 nodes per layer. The acoustic feature is 80-dimension log Mel filter bank for every 10 ms speech.  

The training data contains 30 thousand (K) hours of transcribed Microsoft data, mixed with 8K and 16K HZ sampled data \cite{li2012improving}. All the data are anonymized with personally identifiable information removed.  The general testing set covers different application scenarios including dictation, conversation, 
far-field speech and call center etc., consisting of a total of 6.4 million
(M) words. For the adaptation testing sets, we selected 2 real application tasks with different size of adaptation text data, as well as Librispeech sets for better reference. The data size of the testing sets are listed in table \ref{tab:testingdata}. The model training never observes the data from these adaptation tasks.

The external LM has the same model structure as the vocabulary predictor in the FNT and was trained with text data containing about 1.5 billion words, which includes the transcription of 30k training data mentioned above for C-T and FNT model training. 

We first evaluate the FNT model's accuracy on general testing set with the proposed methods, including adding CTC criterion, initializing the vocabulary predictor from a well trained external LM. Then we examine the performance of above FNT models on adaptation sets by fine-tuning the vocabulary predictor with text adaptation data. Finally, n-gram interpolation adaptation method is evaluated based on the best FNT model from above experiments.   
\begin{table}[t]
    \centering
    \begin{tabular}{c|c|c}
    \hline
       testing set        & adaptation data    & testing data \\
     \hline
      task1       &   6,135   &  6,269    \\
      \hline
      task2       &   193,047   &  21,960    \\
    \hline
      Librispeech &   18,740,565   &  210,246    \\ 
     \hline
    \end{tabular}
    \caption{Word count for testing sets.}
    \vspace{-0.0cm}
    \label{tab:testingdata}
\end{table}
\subsection{Results on general testing set}
\label{ssec:generalset}

All results on general testing set are given in table \ref{tab:generalset}. The baseline model is a standard C-T model. The standard FNT model is with the same structure as in \cite{fnt}, and it is trained with 30k training data from scratch. Compared with the baseline C-T model, the standard FNT model got 1.29\% relative WER increase. To reduce the accuracy degradation, FNT model is refined by adding the CTC criterion based on equation \ref{eqn:ctc} and \ref{eqn:loss}. The CTC loss weight $\beta$ is 0.1. We can see that adding CTC decreases WER from 11.01 to 10.97, but is still worse than the baseline C-T model. The degradation is reduced further by initializing vocabulary predictor from an external language model. Two recipes are examined: one is the external LM is fixed during the FNT model training, and the other is the external LM is updated together with other parts of FNT model. Updating the external LM got the best result, which is even better than the baseline C-T model.

\begin{table}[th]
    \centering
    \begin{tabular}{l|c}
    \hline
          Model        & General set \\
     \hline
      Baseline C-T (B0)      &       10.87 \\
    \hline
      Standard FNT (F0) & 11.01 \\
  \quad+CTC (F1) & 10.97 \\
       \quad\quad +ext. LM fix (F2) & 10.89 \\
       \quad\quad\quad+ext. LM update (F3) & 10.70 \\
     \hline
    \end{tabular}
    \caption{WER(\%) on 
    the general testing set.}
    \vspace{-0.0cm}
    \label{tab:generalset}
\end{table}
\subsection{Results on adaptation testing sets}
\label{ssec:adaptation}

In this section, We first evaluated the impact of KL divergence using Librispeech set based on standard FNT model. The results are given in table \ref{tab:kl}. The results showed the adapted model without KL divergence degraded the accuracy on general testing set largely. FNT adaptation with KL divergence helped to recovered the loss on general testing set obviously with very small WER increase on adaptation set compared to the standard FNT adaptation. In the following adaptation experiments, KL divergence weight $\alpha$ is always set to 0.1.

Table \ref{tab:adaptationset} shows the adaptation results for different FNT models on all adaptation sets. For each task, the results in ``base'' column are WER before adaptation, and the results in ``adapt'' column are WER after adaptation. Simple average WERs are also reported by averaging the WERs from all three tasks. Comparing ``base'' results for B0 and F0, we could find the accuracy gap between the baseline C-T model and the standard FNT model on these adaptation set is much larger than that on general set, especially for task1 and task2. The possible reason is that the domains in the training data may have some coverage for the general testing set, but they are totally irrelevant to these adaptation sets. With the proposed refinements for FNT, this gap is decreased step by step, the best FNT model F3, which is with full combination of the proposed methods, could get the similar accuracy as the baseline C-T model. The same trend could be also observed for the ``adapt'' results. Each method contributes accuracy improvement to the adapted FNT model. Compared with the adapted standard FNT model (F0), the adapted model F3 reduces WER for three adaptation sets by relatively 9.48\%   in average (from 12.80\% to 11.59\% ). And compared with the baseline C-T model, the adapted model F3 gets 29.21\% relative WER reduction (from 16.37\% to 11.59\%).

\newcommand{\specialcell}[2][c]{%
  \begin{tabular}[#1]{@{}c@{}}#2\end{tabular}}
  
\begin{table}[h]
    \centering
    \begin{tabular}{l|c|c|c|c|c}     
      \hline
      \multirow{2}{*}{} & Standard  & \multicolumn{4}{c}{KLD weight} \\\cline{3-6}
                & FNT & 0.0 & 0.1 & 0.2 & 0.3 \\
      \hline
     
      General set      &      10.87  & 12.78 & 11.69 &11.52 & 11.4 \\
      \hline
      Librispeech       &      8.32  &7.17 & 7.17 &7.25 & 7.33 \\
      \hline
    \end{tabular}
    \caption{WER(\%) for Librispeech with different KLD weights.}
    \vspace{-0.0cm}
    \label{tab:kl}
\end{table}



\begin{table}[th]

    \centering
    \resizebox{0.99\columnwidth}{!}{
    \begin{tabular}{l|c|c|c|c|c|c|c|c}
    \hline
          \multirow{2}{*}{Model} & \multicolumn{2}{c}{Task1} & \multicolumn{2}{|c|}{Task2} & \multicolumn{2}{|c|}{Librispeech} & \multicolumn{2}{c}{Simple average}  \\\cline{2-9}
            & Base&Adapt & Base&Adapt & Base&Adapt & Base&Adapt \\
     \hline
      B0     &       16.59 & - &24.19 & - &8.32&- & 16.37 &-\\
    \hline
      F0 & 17.45 & 11.8 & 25.21 & 19.43& 8.44 & 7.17 & 17.03 & 12.80 \\
  F1 & 17.21 & 10.89 & 25.05 & 19.04 & 8.43 &  7.07 & 16.90 & 12.33\\
       F2 & 17.63 & 10.8 & 24.96 & 18.55 & 8.41  &7.16 &17.00 &12.17 \\
       F3 & 16.38 & 10.19 & 24.44 & 17.7 & 8.33 & 6.87&16.38 &11.59 \\
     F3+n-gram & 16.38 & 11.49 & 24.44 & 19.89 & 8.33 & 7.39& 16.38& 12.92 \\
     \hline
    \end{tabular}
    }
    \caption{WER(\%) on 
    adaptation testing sets.  }
    \vspace{-0.0cm}
    \label{tab:adaptationset}
\end{table}

\subsection{Results of n-gram interpolation}
\label{ssec:result_ngram}

In this section, we evaluated n-gram interpolation performance based on the best FNT model (F3) for three adaptation tasks. For each task, a 5-gram LM is trained with the adaptation text. To make the interpolation simple and efficient, sentence piece instead of word is used as the basic unit for the 5-gram LM. The interpolation weight $w$ is always set as 0.3. The results are shown in the last row of table \ref{tab:adaptationset}. Compared with the adaptation of fine-tuning vocabulary predictor, n-gram interpolation based adaptation got a little higher WER, but the relative WER reduction over the baseline C-T model is still satisfying, which is 21.04\% (from 16.37\% to 12.92\%). More importantly, the adaptation speed is improved hugely. Experiments show that for fine-tuning method, the adaptation process cost about 10 seconds per 1,000 words on GPU and the cost is formidable when adapting with CPU. In contrast, it only needs about 0.002 seconds per 1,000 words on CPU for n-gram LM training. This is very useful for those  application scenarios which need immediate adaptation.

\section{conclusions}
\label{sec:conclusion}

In this paper, several methods are proposed to improve the accuracy and efficiency for FNT adaptation with text-only data. These methods include: 1) during the FNT model training, adding CTC criterion to make the encoder act more like an acoustic model and initializing vocabulary predictor with a well trained external LM to use more text data. 2) during the FNT model adaptation, adding KL divergence to avoid over fitting to the adaptation data. 3) using n-gram interpolation with the vocabulary-prediction LM module inside FNT instead of fine-tuning the vocabulary-prediction LM module to improve the adaptation speed. The experimental results proved that, compared with standard FNT, the proposed methods could get better accuracy on general testing set, and decrease the adaptation WER by 9.48\% percent relatively.  In total, compared with the baseline C-T model, the adaptation WER is decreased by 29.21\% relatively. Besides, n-gram interpolation could get much faster adaptation than the fine-tuning method, enabling the scenarios which require immediate adaptation.

\bibliographystyle{IEEEbib}
\bibliography{refs}

\end{document}